\documentclass[11pt,a4paper]{article}
\usepackage[hyperref]{acl2019}
\usepackage{times}
\usepackage{latexsym}

\usepackage{url}
\usepackage{helvet}  
\usepackage{courier}  
\usepackage{graphicx}  
\usepackage{multirow}
\usepackage{xspace}
\usepackage{amssymb}
\usepackage{amsmath}
\usepackage{subfigure}
\usepackage{booktabs}
\usepackage{bbm}
\usepackage{array}
\usepackage{bm}
\usepackage[noend]{algpseudocode}
\usepackage{algorithmicx,algorithm}
\usepackage{amssymb}
\usepackage{pifont}
\DeclareMathOperator*{\argmax}{arg\,max}
\graphicspath{{figures/}} 

\aclfinalcopy 

\newcommand\bertbase{BERT$_{\textsc{BASE}}$\xspace}
\newcommand\bertlarge{BERT$_{\textsc{LARGE}}$\xspace}
\newcommand\bertpipe{BERT$_{\textsc{pipe}}$\xspace}
\newcommand\bertpipeplus{BERT$_{\textsc{pipe}}$*\xspace}
\newcommand\reqabase{RE$^3$QA$_{\textsc{BASE}}$\xspace}
\newcommand\reqalarge{RE$^3$QA$_{\textsc{LARGE}}$\xspace}
\newcommand{\tabincell}[2]{\begin{tabular}{@{}#1@{}}#2\end{tabular}}
\newcommand{\cmark}{\ding{51}}%
\newcommand{\xmark}{\ding{55}}%

\title{Retrieve, Read, Rerank: Towards End-to-End \\ Multi-Document Reading Comprehension}

\author{
Minghao Hu,
Yuxing Peng,
Zhen Huang,
Dongsheng Li \\
National University of Defense Technology, Changsha, China \\
{\tt \{huminghao09,pengyuxing,huangzhen,dsli\}@nudt.edu.cn}
}

\date{}

\begin{document}
\maketitle

\begin{abstract}

This paper considers the reading comprehension task in which multiple documents are given as input.
Prior work has shown that a pipeline of retriever, reader, and reranker can improve the overall performance.
However, the pipeline system is inefficient since the input is re-encoded within each module, and is unable to leverage upstream components to help downstream training.
In this work, we present RE$^3$QA, a unified question answering model that combines context retrieving, reading comprehension, and answer reranking to predict the final answer.
Unlike previous pipelined approaches, RE$^3$QA shares contextualized text representation across different components, and is carefully designed to use high-quality upstream outputs (e.g., retrieved context or candidate answers) for directly supervising downstream modules (e.g., the reader or the reranker).
As a result, the whole network can be trained end-to-end to avoid the context inconsistency problem.
Experiments show that our model outperforms the pipelined baseline and achieves state-of-the-art results on two versions of TriviaQA and two variants of SQuAD.

\end{abstract}
\section{Introduction}
Teaching machines to read and comprehend text is a long-term goal of natural language processing.
Despite recent success in leveraging reading comprehension (RC) models to answer questions given a related paragraph~\cite{wang2017gated,hu2017reinforced,Yu18}, extracting answers from documents or even a large corpus of text (e.g., Wikipedia or the whole web) remains to be an open challenge.
This paper considers the multi-document RC task~\cite{joshi2017triviaqa}, where the system needs to, given a question, identify the answer from multiple evidence documents.
Unlike single-pargraph settings~\cite{Rajpurkar16}, this task typically involves a \emph{retriever} for selecting few relevant document content~\cite{chen2017reading}, a \emph{reader} for extracting answers from the retrieved context~\cite{clark2017simple}, and even a \emph{reranker} for rescoring multiple candidate answers~\cite{bogdanova2016we}.

Previous approaches such as DS-QA~\cite{lin2018denoising} and R$^3$~\cite{wang2018r3} consist of separate retriever and reader models that are jointly trained.
\citet{wang2018joint} further propose to rerank multiple candidates for verifying the final answer.
\citet{wang2017evidence} investigate the full retrieve-read-rerank process by constructing a pipeline system that combines an information retrieval (IR) engine, a neural reader, and two kinds of answer rerankers.
Nevertheless, the pipeline system requires re-encoding inputs for each subtask, which is inefficient for large RC tasks. 
Moreover, as each model is trained independently, high-quality upstream outputs can not benefit downstream modules.
For example, as the training proceeds, a neural retriever is able to provide more relevant context than an IR engine~\cite{htut2018training}. 
However, the reader is still trained on the initial context retrieved using IR techniques.
As a result, the reader could face a \emph{context inconsistency} problem once the neural retriever is used.
Similar observation has been made by~\citet{wang2018multi}, where integrating both the reader and the reranker into a unified network is more benefical than a pipeline (see Table \ref{table:model} for more details).

\begin{table*}
\begin{center}
\begin{tabular}{l|c|c|c|c}
\toprule
Model & Retrieve & Read & Rerank & Architecture  \\ 
\midrule	
DS-QA~\cite{lin2018denoising}       & \cmark & \cmark & \xmark & Pipeline \\
R$^3$~\cite{wang2018r3}             & \cmark & \cmark & \xmark & Pipeline* \\
Extract-Select~\cite{wang2018joint} & \xmark & \cmark & \cmark & Pipeline* \\
V-Net~\cite{wang2018multi}          & \xmark & \cmark & \cmark & Unified  \\
Re-Ranker~\cite{wang2017evidence}   & \cmark & \cmark & \cmark & Pipeline \\
\textbf{RE$^3$QA}                   & \cmark & \cmark & \cmark & Unified  \\
\bottomrule
\end{tabular}
\caption{\label{table:model} Comparison of RE$^3$QA with existing approaches.
Our approach performs the full retrieve-read-rerank process with a unified network instead of a pipeline of separate models. *: R$^3$ and Extract-Select jointly train two models with reinforcement learning.}
\end{center}
\end{table*}

In this paper, we propose RE$^3$QA, a neural question answering model that conducts the full \textbf{re}trieve-\textbf{re}ad-\textbf{re}rank process for multi-document RC tasks.
Unlike previous pipelined approaches that contain separate models, we integrate an early-stopped retriever, a distantly-supervised reader, and a span-level answer reranker into a unified network.
Specifically, we encode segments of text with pre-trained Transformer blocks~\cite{devlin2018bert}, where earlier blocks are used to predict retrieving scores and later blocks are fed with few top-ranked segments to produce multiple candidate answers.
Redundant candidates are pruned and the rest are reranked using their span representations extracted from the shared contextualized representation.
The final answer is chosen according to three factors: the retrieving, reading, and reranking scores.
The whole network is trained end-to-end so that the context inconsistency problem can be alleviated.
Besides, we can avoid re-encoding input segments by sharing contextualized representations across different components, thus achieving better efficiency.

We evaluate our approach on four datasets.
On TriviaQA-Wikipedia and TriviaQA-unfiltered datasets~\cite{joshi2017triviaqa}, we achieve 75.2 F1 and 71.2 F1 respectively, outperforming previous best approaches.
On SQuAD-document and SQuAD-open datasets, both of which are modified versions of SQuAD~\cite{Rajpurkar16}, we obtain 14.8 and 4.1 absolute gains on F1 score over prior state-of-the-art results. 
Moreover, our approach surpasses the pipelined baseline with faster inference speed on both TriviaQA-Wikipedia and SQuAD-document.
Source code is released for future research exploration\footnote{https://github.com/huminghao16/RE3QA}.
\section{Related Work}
Recently, several large datasets have been proposed to facilitate the research in document-level reading comprehension (RC)~\cite{clark2017simple} or even open-domain question answering~\cite{chen2017reading}.
TriviaQA~\cite{joshi2017triviaqa} is a challenging dataset containing over 650K question-answer-document triples, in which the document are either Wikipedia articles or web pages.
Quasar-T~\cite{dhingra2017quasar} and SearchQA~\cite{dunn2017searchqa}, however, pair each question-answer pair with a set of web page snippets that are more analogous to paragraphs.
Since this paper considers the multi-document RC task, we therefore choose to work on TriviaQA and two variants of SQuAD~\cite{Rajpurkar16}.

To tackle this task, previous approaches typically first retrieve relevant document content and then extract answers from the retrieved context.
\citet{choi2017coarse} construct a coarse-to-fine framework that answers the question from a retrieved document summary.
\citet{wang2018r3} jointly train a ranker and a reader with reinforcement learning~\cite{sutton2011reinforcement}.
\citet{lin2018denoising} propose a pipeline system consisting of a paragraph selector and a paragraph reader.
\citet{yang2019end} combine BERT with an IR toolkit for open-domain question answering.

However, \citet{Jia17} show that the RC models are easily fooled by adversarial examples.
By only extracting an answer without verifying it, the models may predict a wrong answer and are unable to recover from such mistakes~\cite{hu2018read}.
In response, \citet{wang2018joint} present an extract-then-select framework that involves candidate extraction and answer selection.
\citet{wang2018multi} introduce a unified network for cross-passage answer verification.
\citet{wang2017evidence} explore two kinds of answer rerankers in an existing retrieve-read pipeline system.
There are some other works that handle this task in different perspectives, such as using hierarchical answer span representations~\cite{pang2018has}, modeling the interaction between the retriever and the reader~\cite{das2018multi}, and so on.

\begin{figure*}
\center
\includegraphics[width=0.95\textwidth]{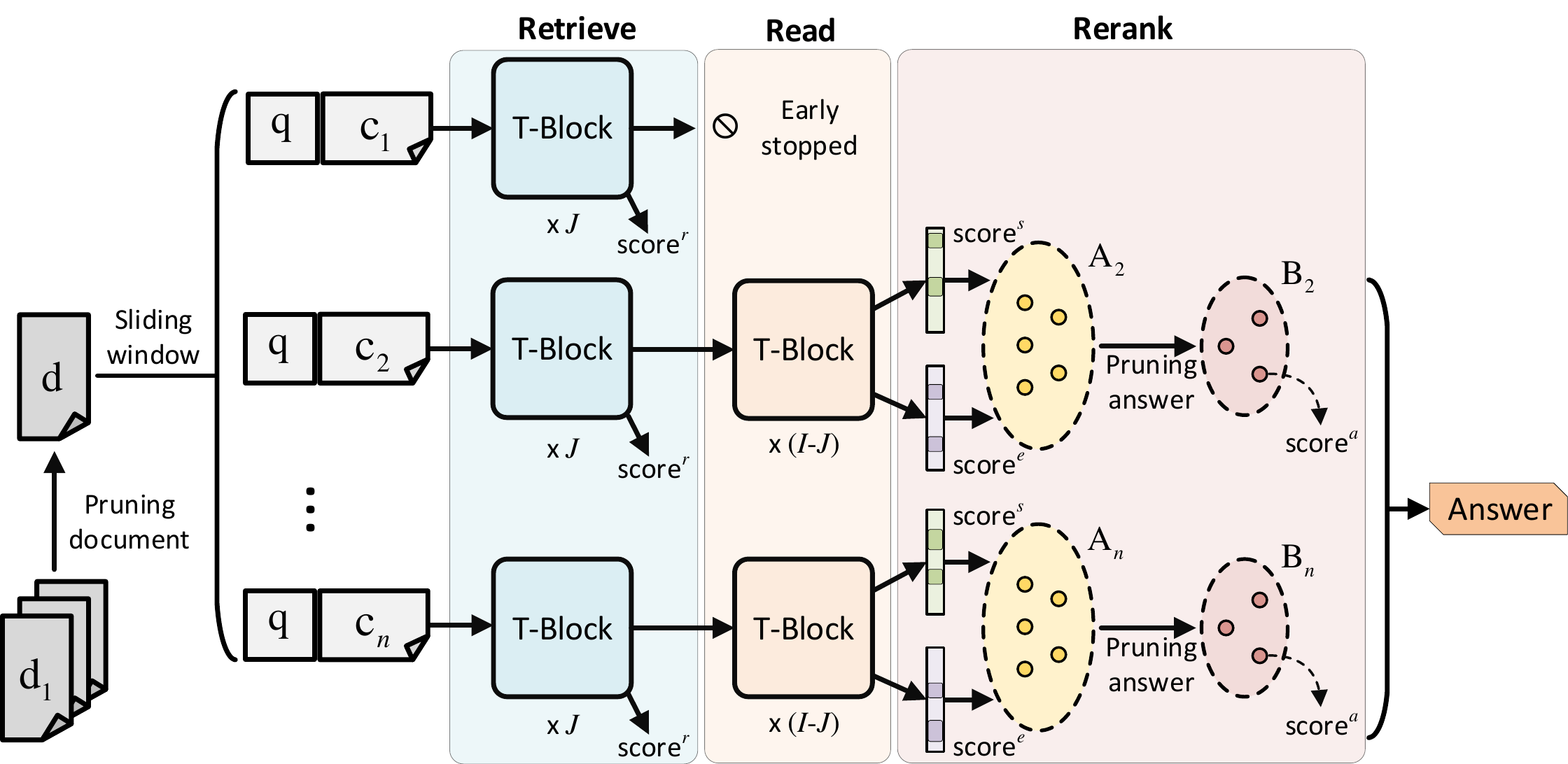}
\caption{RE$^3$QA architecture. 
The input documents are pruned and splitted into multiple segments of text, which are then fed into the model\footnote{blabla}. 
Few top-ranked segments are retrieved and the rest are early stopped. 
Multiple candidate answers are proposed for each segment, which are later pruned and reranked.
RE$^3$QA has three outputs per candidate answer: the retrieving, reading, and reranking scores.
The network is trained end-to-end with a multi-task objective.
``T-Block'' refers to pre-trained Transformer block~\cite{devlin2018bert}. }
\label{fig:re3qa_overview}
\end{figure*}

Our model differs from these approaches in several ways: 
(a) we integrate the retriever, reader, and reranker components into a unified network instead of a pipeline of separate models, 
(b) we share contextualized representation across different components while pipelined approaches re-encode inputs for each model,   
and (c) we propose an end-to-end training strategy so that the context inconsistency problem can be alleviated.

A cascaded approach is recently proposed by \citet{yan2018deep}, which also combines several components such as the retriever and the reader while sharing several sets of parameters.
Our approach is different in that we ignore the document retrieval step since a minimal context phenomenon has been observed by \citet{min2018efficient}, and we additionally consider answer reranking.

\section{RE$^3$QA}
Figure \ref{fig:re3qa_overview} gives an overview of our multi-document reading comprehension approach.
Formally, given a question and a set of documents, we first filter out irrelevant document content to narrow the search space (\S\ref{sec:doc-prun}). 
We then split the remaining context into multiple overlapping, fixed-length text segments. 
Next, we encode these segments along with the question using pre-trained Transformer blocks~\cite{devlin2018bert} (\S\ref{sec:seg-enc}).
To maintain efficiency, the model computes a retrieving score based on shallow contextual representations with early summarization, and only returns a few top-ranked segments (\S\ref{sec:seg-retri}). 
It then continues encoding these retrieved segments and outputs multiple candidate answers under the distant supervision setting (\S\ref{sec:seg-read}).
Finally, redundant candidates are pruned and the rest are reranked using their span representations (\S\ref{sec:ans-rerank}). 
The final answer is chosen according to the retrieving, reading, and reranking scores.
Our model is trained end-to-end\footnote{Note that ``end-to-end training'' only involves retrieving, reading, and reranking, but not the very first pruning step.} by back-propagation (\S\ref{sec:train}).

\subsection{Document Pruning	\label{sec:doc-prun}}
The input to our model is a question $\mathbf{q}$ and a set of documents $\mathbf{D}=\{\mathbf{d}_1, ..., \mathbf{d}_{N_D}\}$. 
Since the documents could be retrieved by a search engine (e.g., up to 50 webpages in the unfiltered version of TriviaQA~\cite{joshi2017triviaqa}) or Wikipedia articles could contain hundreds of paragraphs, we therefore first discard irrelevant document content at paragraph level. 
Following ~\citet{clark2017simple}, we select the top-$K$ paragraphs that have smallest TF-IDF cosine distances with each question. 
These paragraphs are then sorted according to their positions in the documents and concatenated to form a new pruned document $\mathbf{d}$.
As a result, a large amount of unrelated text can be filtered out while a high recall is guaranteed.
For example, nearly 95\% of context are discarded while the chance of selected paragraphs containing correct answers is 84.3\% in TriviaQA-unfiltered.

\subsection{Segment Encoding \label{sec:seg-enc}}
Typically, existing approaches either read the retrieved document at the paragraph level~\cite{clark2017simple} or at the sentence level~\cite{min2018efficient}. 
Instead, following~\citet{hewlett2017accurate}, we slide a window of length $l$ with a stride $r$ over the pruned document $\mathbf{d}$ and produce a set of text segments $\mathbf{C}=\{\mathbf{c}_1, ..., \mathbf{c}_n\}$, where $n= \left \lceil{\frac{L_d - l}{r}}\right \rceil +1$, and $L_d$ is the document length. 
Next, we encode these segments along with the question using pre-trained Transformer blocks~\cite{devlin2018bert}, which is a highly parallel encoding scheme instead of recurrent approaches such as LSTMs.

The input to the network is a sequence of tokens $\mathbf{x}=(x_1, ..., x_{L_x})$ with length $L_x$.
It is obtained by concatenating the question, segment, and several delimiters as $[\texttt{[CLS]}; \mathbf{q}; \texttt{[SEP]}; \mathbf{c}; \texttt{[SEP]}]$, where \texttt{[CLS]} is a classification token and \texttt{[SEP]} is another token for differentiating sentences.
We refer to this sequence as ``segment'' in the rest of this paper.
For each token $x_i$ in $\mathbf{x}$, its input representation is the element-wise addition of word, type, and position embeddings.
Then, we can obtain the input embeddings $\mathbf{h}^0 \in\mathbb{R}^{L_x \times D_h}$, where $D_h$ is hidden size.

Next, a series of $I$ pre-trained Transformer blocks are used to project the input embeddings into a sequence of contextualized vectors as:
\begin{equation}
\mathbf{h}^i = \mathrm{TransformerBlock}(\mathbf{h}^{i-1}), \forall i \in [1,I] \nonumber
\end{equation}
Here, we omit a detailed introduction on the block architecture and refer readers to \citet{vaswani2017attention} for more details.

\subsection{Early-Stopped Retriever \label{sec:seg-retri}}
While we find the above parallel encoding scheme very appealing, there is a crucial computational inefficiency if all segments are fully encoded. 
For example, the average number of segments per instance in TriviaQA-unfiltered is 20 even after pruning, while the total number of Transformer blocks is 12 or 24.
Therefore, we propose to rank all segments using early-summarized hidden representations as a mechanism for efficiently retrieving few top-ranked segments.

Specifically, let $\mathbf{h}^J$ denote the hidden states in the $J$-th block, where $J < I$. 
We compute a $\mathbf{score}^r \in\mathbb{R}^{2}$ by summarizing $\mathbf{h}^J$ into a fix-sized vector with a weighted self aligning layer followed by multi-layer perceptrons as:
\begin{gather}
\mu = \mathrm{softmax}(\mathbf{w}_{\mu} \mathbf{h}^J) \nonumber \\
\mathbf{score}^r = \mathbf{w}_r \mathrm{tanh}(\mathbf{W}_r  \sum\nolimits_{i=1}^{L_x} {\mu_i} \mathbf{h}_i^J) \nonumber  
\end{gather}
where $\mathbf{w}_{\mu}$, $\mathbf{w}_r$, $\mathbf{W}_r$ are parameters to be learned. 

After obtaining the scores of all segments, we pass the top-$N$ ranked segments per instance to the subsequent blocks, and discard the rest. 
Here, $N$ is relatively small so that the model can focus on reading the most revelant context.

To train the retrieving component, we normalize $\mathbf{score}^r$ and define the objective function as:
\begin{equation} \label{eq:1}
	\mathcal{L}_{\uppercase\expandafter{\romannumeral1}} = - \sum\nolimits_{i=1}^{2} \mathbf{y}^r_i \log ( \mathrm{softmax}(\mathbf{score}^r)_i )
\end{equation}
where $\mathbf{y}^r$ is an one-hot label indicating whether current segment contains at least one exactly-matched ground truth answer text or not.

\subsection{Distantly-Supervised Reader \label{sec:seg-read}}
Given the retrieved segments, the reading component aims to propose multiple candidate answers per segment. 
This is achieved by first element-wisely projecting the final hidden states $\mathbf{h}^I$ into two sets of scores as follows:
\begin{equation}
\mathbf{score}^s = \mathbf{w}_s \mathbf{h}^I \ , \  \mathbf{score}^e = \mathbf{w}_e \mathbf{h}^I \nonumber
\end{equation}
where $\mathbf{score}^s \in\mathbb{R}^{L_x}$ and $\mathbf{score}^e \in\mathbb{R}^{L_x}$ are the scores for the start and end positions of answer spans, and $\mathbf{w}_s$, $\mathbf{w}_e$ are trainable parameter vectors.

Next, let $\alpha_i$ and $\beta_i$ denote the start and end indices of candidate answer $\mathbf{a}_i$. 
We compute a reading score, $\mathbf{s}_i = \mathbf{score}^s_{\alpha_i} + \mathbf{score}^e_{\beta_i}$, and then propose top-$M$ candidates according to the descending order of the scores, yielding a set of preliminary candidate answers $\mathbf{A}=\{\mathbf{a}_1, ..., \mathbf{a}_M\}$ along with their scores $\mathbf{S}=\{\mathbf{s}_1, ..., \mathbf{s}_M\}$.

Following previous work~\cite{clark2017simple}, we label all text spans within a segment that match the gold answer as being correct, thus yielding two label vectors $\mathbf{y}^s \in\mathbb{R}^{L_x}$ and $\mathbf{y}^e \in\mathbb{R}^{L_x}$. 
Since there is a chance that the segment does not contain any answer string, we then label the first element in both $\mathbf{y}^s$ and $\mathbf{y}^e$ as 1, and set the rest as 0.
Finally, we define the objective function as:
\begin{align} \label{eq:2}
	\mathcal{L}_{\uppercase\expandafter{\romannumeral2}} = & - \sum\nolimits_{i=1}^{L_x} \mathbf{y}^s_i \log (\mathrm{softmax}(\mathbf{score}^s)_i)	\nonumber \\
	& - \sum\nolimits_{j=1}^{L_x} \mathbf{y}^e_j \log (\mathrm{softmax}(\mathbf{score}^e)_j)
\end{align}

\subsection{Answer Reranker \label{sec:ans-rerank}}
The answer reranker aims to rerank the candidate answers proposed by the previous reader.
We first introduce a span-level non-maximum suppression algorithm to prune redundant candidate spans, and then predict the reranking scores for remaining candidates using their span representations.

\paragraph{Span-level non-maximum suppression}
So far, the reader has proposed multiple candidate spans.
However, since there is no constraint to predict an unique span for an answer string, multiple candidates may refer to the same text.
As a result, other than the first correct span, all other spans on the same text would be false positives.
Figure \ref{fig:span_nms} shows a qualitative example of this phenomenon.

\begin{figure}[h]
\center
\fbox{\parbox{0.95\columnwidth}{
\begin{small}
\textbf{Question:} In the late 60s Owen Finlay MacLaren pioneered what useful item for parents of small chldren?

\textbf{Answer:} baby buggy

\textbf{Candidates:}
baby buggy, collapsible baby buggy, buggy, folding buggy, folding chair ...
\end{small}
}}
\caption{An example from TriviaQA shows that multiple candidate answers refer to the same text.}
\label{fig:span_nms}
\end{figure}

Inspired by the non-maximum suppression (NMS) algorithm~\cite{rosenfeld1971edge} that is used to prune redundant bounding boxes in object detection~\cite{ren2015faster}, we present a span-level NMS (Algorithm \ref{algorithm1}) to alleviate the problem.
Specifically, span-level NMS starts with a set of candidate answers $\mathbf{A}$ with scores $\mathbf{S}$.
After selecting the answer $\mathbf{a}_i$ that possesses the maximum score, we remove it from the set $\mathbf{A}$ and add it to $\mathbf{B}$. 
We also delete any answer $\mathbf{a}_j$ in $\mathbf{A}$ that is overlapped with $\mathbf{a}_i$.
We define that two candidates overlap with each other if they share at least one boundary position\footnote{We also experimented with the span-level F1 function, but found no performance improment.}.
This process is repeated for remaining answers in $\mathbf{A}$, until $\mathbf{A}$ is empty or the size of $\mathbf{B}$ reaches a maximum threshold.

\begin{algorithm}[h]
\small
\caption{Span-level NMS} 
\label{algorithm1}
{\bf Input:} 
$\mathbf{A}=\{\mathbf{a}_i\}_{i=1}^M$; $\mathbf{S}=\{\mathbf{s}_i\}_{i=1}^M$; $M^*$ \\
\hspace*{0.15in} $\mathbf{A}$ is the set of preliminary candidate answers \\
\hspace*{0.15in} $\mathbf{S}$ is the corresponding confidence scores \\
\hspace*{0.15in} $M^*$ denotes the maximum size threshold
\begin{algorithmic}[1]
\State Initialize $\mathbf{B} = \{\}$ 

\While{$\mathbf{A} \ne \{\}$ and $\mathrm{size}(\mathbf{B}) < M^*$} 
　　\State $i = \argmax \mathbf{S}$
	\State $\mathbf{B} = \mathbf{B} \cup \{\mathbf{a}_i$\}; $\mathbf{A} = \mathbf{A} - \{\mathbf{a}_i$\}; $\mathbf{S} = \mathbf{S} - \{\mathbf{s}_i\}$

	\For{$\mathbf{a}_j$ in $\mathbf{A}$} 
	　　\If{$\mathrm{overlap}(\mathbf{a}_i, \mathbf{a}_j)$} 
	　　　　\State $\mathbf{A} = \mathbf{A} - \{\mathbf{a}_j\}$; $\mathbf{S} = \mathbf{S} - \{\mathbf{s}_j\}$
	　　\EndIf
	\EndFor

\EndWhile
\State \Return $\mathbf{B}$
\end{algorithmic}
\end{algorithm}

\paragraph{Candidate answer reranking}
Given the candidate answer $\mathbf{a}_i$ in $\mathbf{B}$, we compute a reranking score based on its span representation, where the representation is a weighted self-aligned vector bounded by the span boundary of the answer, similar to \citet{lee2017end,he2018jointly}:
\begin{gather}
\eta = \mathrm{softmax}(\mathbf{w}_{\eta} \mathbf{h}^I_{\alpha_i:\beta_i}) \nonumber \\
\mathbf{score}^a_i = \mathbf{w}_a \mathrm{tanh}(\mathbf{W}_a  \sum\nolimits_{j=\alpha_i}^{\beta_i} {\eta_{j-\alpha_i+1}} \mathbf{h}_j^I)  \nonumber
\end{gather}
Here, $\mathbf{score}^a \in\mathbb{R}^{M^*}$, and $\mathbf{h}^I_{\alpha_i:\beta_i}$ is a shorthand for stacking a list of vectors $\mathbf{h}^I_j$ ($\alpha_i \le j \le \beta_i$).

To train the reranker, we construct two kinds of labels for each candidate $\mathbf{a}_i$.
First, we define a hard label $\mathbf{y}^{hard}_i$ as the maximum exact match score between $\mathbf{a}_i$ and ground truth answers.
Second, we also utilize a soft label $\mathbf{y}^{soft}_i$, which is computed as the maximum F1 score between $\mathbf{a}_i$ and gold answers, so that the partially correct prediction can still have a supervised signal. 
The above labels are annotated for each candidate in $\mathbf{B}$, yielding $\mathbf{y}^{hard} \in\mathbb{R}^{M^*}$ and $\mathbf{y}^{soft} \in\mathbb{R}^{M^*}$.
If there is no correct prediction in $\mathbf{B}$ (all elements of $\mathbf{y}^{hard}$ are 0), then we replace the least confident candidate with a gold answer.
Finally, we define the following reranking objective:
\begin{align}  \label{eq:3}
	\mathcal{L}_{\uppercase\expandafter{\romannumeral3}} & = - \sum\nolimits_{i=1}^{M^*} \mathbf{y}^{hard}_i \log (\mathrm{softmax}(\mathbf{score}^a)_i)	\nonumber \\
	& + \sum\nolimits_{i=1}^{M^*} ||\mathbf{y}^{soft}_i - \frac{\mathbf{score}^a_i}{ \sum\nolimits_{j=1}^{M^*}\mathbf{score}^a_j }||^2
\end{align}

\begin{table*}
\begin{center}
\begin{tabular}{l|ccccccccc}
\toprule
Dataset  & \#Ins  & \#Doc & \#Seg & \#Tok & \#Tok* & $K$ & $N$ & Recall \\ 
\midrule
TriviaQA-Wikipedia     & 7,993  & 1.8  & 17 & 10,256 & 2,103 & 14 & 8 & 94.8\% \\
TriviaQA-unfiltered    & 11,313 & 11.7 & 20 & 52,635 & 2,542 & 14 & 8 & 84.3\% \\
SQuAD-document         & 10,570 & 1    & 35 & 5,287  & 3,666 & 30 & 8 & 99.0\% \\
SQuAD-open             & 10,570 & 5    & 42 & 38,159 & 5,103 & 30 & 8 & 64.9\% \\
\bottomrule
\end{tabular}
\caption{\label{table:data} Dataset statistics. `\#Ins' denotes the number of instances, while `\#Doc', `\#Seg', `\#Tok', and `\#Tok*' refer to the average number of documents, segments, and tokens before/after pruning, respectively. $K$ and $N$ are the number of retrieved paragraphs and segments. All statistics are calculated on the development set.}
\end{center}
\end{table*}

\subsection{Training and Inference \label{sec:train}}
Rather than separately training each component, we propose an end-to-end training strategy so that downstream components (e.g., the reader) can benefit from the high-quality upstream outputs (e.g., the retrieved segments) during training.

Specifically, we take a multi-task learning approach~\cite{caruna1993multitask,ruder2017overview}, sharing the parameters of earlier blocks with a joint objective function defined as:
\begin{equation} 
	\mathcal{J} = \mathcal{L}_{\uppercase\expandafter{\romannumeral1}} + \mathcal{L}_{\uppercase\expandafter{\romannumeral2}} + \mathcal{L}_{\uppercase\expandafter{\romannumeral3}}  	\nonumber
\end{equation}

Algorithm \ref{algorithm2} details the training process.
Before each epoch, we compute $\mathbf{score}^r$ for all segments in the training set $\mathcal{X}$.
Then, we retrieve top-$N$ segments per instance and construct a new training set $\mathcal{\tilde{X}}$, which only contains retrieved segments.
For each instance, if all of its top-ranked segments are negative examples, then we replace the least confident one with a gold segment.
During each epoch, we sample two sets of mini-batch from both the $\mathcal{X}$ and the $\mathcal{\tilde{X}}$, where the first batch is used to calculate $\mathcal{L}_{\uppercase\expandafter{\romannumeral1}}$ and the other one for computing $\mathcal{L}_{\uppercase\expandafter{\romannumeral2}}$ and $\mathcal{L}_{\uppercase\expandafter{\romannumeral3}}$.
Note that the contextualized vectors $\mathbf{h}^I$ are shared across the reader and the reranker to avoid repeated computations.
The batch size of $\mathcal{X}$ is dynamically decided so that both of $\mathcal{X}$ and $\mathcal{\tilde{X}}$ can be traversed with the same number of steps.

During inference, we take the retrieving, reading, and reranking scores into account.
We compare the scores across all segments from the same instance, and choose the final answer according to the weighted addition of these three scores.

\begin{algorithm}[h]
\small
\caption{End-to-end training of RE$^3$QA} 
\label{algorithm2}
{\bf Input:} 
$\mathcal{X}=\{\mathbf{X}_i\}_{i=1}^t$, where $\mathbf{X}_i=\{\mathbf{x}_i^j\}_{j=1}^n$; $\mathbf{M}_{\Theta}$; $k$ \\
\hspace*{0.15in} $\mathcal{X}$ is the dataset containing $t$ instances \\
\hspace*{0.15in} $\mathbf{X}^i$ is $i$-th instance containing $n$ segments \\
\hspace*{0.15in} $\mathbf{M}_{\Theta}$ denotes the model with parameters $\Theta$ \\
\hspace*{0.15in} $k$ is the maximum number of epoch
\begin{algorithmic}[1]
\State Initialize $\Theta$ from pre-trained parameters 

\For{$\mathtt{epoch}$ in $1,...,k$}
	\State Compute $\mathbf{score}^r$ for all $\mathbf{x}$ in $\mathcal{X}$
	\State Retrieve top-$N$ segments per instance
	\State Construct a new $\mathcal{\tilde{X}}$ that includes retrieved $\mathbf{x}$

	\For{$\mathtt{batch}_{\mathcal{X}}$, $\mathtt{batch}_{\mathcal{\tilde{X}}}$ in $\mathcal{X}$, $\mathcal{\tilde{X}}$}
		\State Compute $\mathcal{L}_{\uppercase\expandafter{\romannumeral1}}$ using $\mathtt{batch}_{\mathcal{X}}$ by Eq. \ref{eq:1}
		\State Compute $\mathcal{L}_{\uppercase\expandafter{\romannumeral2}}$ using $\mathtt{batch}_{\mathcal{\tilde{X}}}$ by Eq. \ref{eq:2}
		\State Reuse $\mathbf{h}^I$ to compute $\mathcal{L}_{\uppercase\expandafter{\romannumeral3}}$ by Eq. \ref{eq:3}
		\State Update $\mathbf{M}_{\Theta}$ with graident $\nabla(\mathcal{J})$
	\EndFor
\EndFor
\end{algorithmic}
\end{algorithm}

\section{Experimental Setup}

\paragraph{Datasets}
We experiment on four datasets: 
(a) TriviaQA-Wikipedia~\cite{joshi2017triviaqa}, a dataset of 77K trivia questions where each question is paired with one or multiple Wikipedia articles.
(b) TriviaQA-unfiltered is a open-domain dataset that contains 99K question-answer tuples. 
The evidence documents are constructed by completing a web search given the question.
(c) SQuAD-document, a variant of SQuAD dataset~\cite{Rajpurkar16} that pairs each question with a full Wikipedia article instead of a specific paragraph.
(d) SQuAD-open~\cite{chen2017reading} is the open domain version of SQuAD where the evidence corpus is the entire Wikipedia domain.
For fair comparision to other methods, we retrieve top-5 articles as our input documents.
The detailed statistics of these datasets are shown in Table \ref{table:data}.

\paragraph{Data preprocessing}
Following~\citet{clark2017simple}, we merge small paragraphs into a single paragraph of up to a threshold length in TriviaQA and SQuAD-open. 
The threshold is set as 200 by default. 
We manually tune the number of retrieved paragraphs $K$ for each dataset, and set the number of retrieved segments $N$ as 8.
Following ~\citet{devlin2018bert}, we set the window length $l$ as $384-L_q-3$ so that $L_x$ is 384 and set the stride $r$ as 128, where $L_q$ is the question length.
We also calculate the answer recall after document pruning, which indicates the performance upper bound.

\paragraph{Model settings}
We initialize our model using two publicly available uncased versions of BERT\footnote{https://github.com/google-research/bert}: \bertbase and \bertlarge, and refer readers to~\citet{devlin2018bert} for details on model sizes.
We use Adam optimizer with a learning rate of 3e-5 and warmup over the first 10\% steps to fine-tune the network for 2 epochs.
The batch size is 32 and a dropout probability of 0.1 is used.
The number of blocks $J$ used for early-stopped retriever is 3 for base model and 6 for large model by default.
The number of proposed answers $M$ is 20, while the threshold of NMS $M^*$ is 5.
During inference, we tune the weights for retrieving, reading, and reranking, and set them as 1.4, 1, 1.4.

\paragraph{Evaluation metrics}
We use mean average precision (MAP) and top-$N$ to evaluate the retrieving component. 
As for evaluating the performance of reading and reranking, we measure the exact match (EM) accuracy and F1 score calculated between the final prediction and gold answers.

\paragraph{Baselines}
We construct two pipelined baselines (denoted as \bertpipe and \bertpipeplus) to investigate the context inconsistency problem.
Both systems contain exactly the same components (e.g., retriever, reader, and reranker) as ours, except that they are trained separately.
For \bertpipe, the reader is trained on the context retrieved by an IR engine. 
As for \bertpipeplus, the reading context is obtained using the trained neural retriever.
\section{Evaluation}
\begin{table}
\begin{center}
\small
\begin{tabular}{l|cccc}
\toprule
\multirow{2}*{ Model } & \multicolumn{2}{c}{ Full } & \multicolumn{2}{c}{ Verified } \\
 & EM & F1 & EM & F1 \\ 
\midrule
\midrule
Baseline$^1$                & 40.3 & 45.9 & 44.9 & 50.7 \\ 
M-Reader$^2$                & 46.9 & 52.9 & 54.5 & 59.5 \\
Re-Ranker$^3$               & 50.2 & 55.5 & 58.7 & 63.2 \\
DrQA$^4$ 					& 52.6 & 58.2 & 57.4 & 62.6 \\
S-Norm$^5$                  & 64.0 & 68.9 & 68.0 & 72.9 \\
MemoReader$^6$              & 64.4 & 69.6 & 70.2 & 75.5 \\
Reading Twice$^7$           & 64.6 & 69.9  & 72.8 & 77.4 \\
SLQA$^8$                    & 66.6 & 71.4 & 74.8 & 78.7 \\
CAPE$^\dagger$				& 67.3 & 72.4 & 75.7 & 79.3 \\
\midrule
\reqabase                 & 68.4 & 72.6 & 76.7 & 79.9 \\
\reqalarge                & \textbf{71.0} & \textbf{75.2} & \textbf{80.3} & \textbf{83.0} \\
\bottomrule
\end{tabular}
\caption{\label{table:tri-wiki} Results on the TriviaQA-Wikipedia test set: \citet{joshi2017triviaqa}$^1$, \citet{hu2017reinforced}$^2$, \citet{wang2017evidence}$^3$, \citet{chen2017reading}$^4$, \citet{clark2017simple}$^5$, \citet{back2018memoreader}$^6$, \citet{weissenborn2017dynamic}$^7$, and \citet{yan2018deep}$^8$. $\dagger$ indicates unpublished works.}
\end{center}
\end{table}

\begin{table}
\begin{center}
\small
\begin{tabular}{l|cccc}
\toprule
Model & EM & F1 \\
\midrule
\midrule
S-Norm~\cite{clark2017simple}               & 64.08 & 72.37 \\ 
\midrule
\reqabase            & 77.90 & 84.81 \\
\reqalarge           & \textbf{80.71} & \textbf{87.20} \\
\bottomrule
\end{tabular}
\caption{\label{table:squad-doc} Results on the SQuAD-document dev set.}
\end{center}
\end{table}

\subsection{Main Results}
Table \ref{table:tri-wiki} summarizes the results on the test set of TriviaQA-Wikipedia dataset. 
As we can see, our best model achieves 71.0 EM and 75.2 F1, firmly outperforming previous methods.
Besides, ~\citet{joshi2017triviaqa} show that the evidence documents contain answers for only 79.7\% of questions in the Wikipedia domain, suggesting that we are approaching the ceiling performance of this task.
However, the score of 80.3 EM on the verified set implies that there is still room for improvement.

We also report the performance on document-level SQuAD in Table \ref{table:squad-doc} to assess our approach in single-document setting. 
We find our approach adapts well: the best model achieves 87.2 F1.
Note that the \bertlarge model has obtained 90.9 F1 on the original SQuAD dataset (single-paragraph setting), which is only 3.7\% ahead of us.

Finally, to validate our approach in open-domain scenarios, we run experiments on the TriviaQA-unfiltered and SQuAD-open datasets, as shown in Table \ref{table:tri-unfil}. 
Again, RE$^3$QA surpasses prior works by an evident margin: our best model achieves 71.2 F1 on TriviaQA-unfiltered, and outperforms a BERT baseline by 4 F1 on SQuAD-open, indicating that our approach is effective for the challenging multi-document RC task.

\begin{table}
\begin{center}
\small
\begin{tabular}{l|cccc}
\toprule
\multirow{2}*{ Model } & \multicolumn{2}{c}{ TriviaQA-unfiltered } & \multicolumn{2}{c}{ SQuAD-open } \\
 & EM & F1 & EM & F1 \\ 
\midrule
\midrule
DrQA$^1$                 & 32.3 & 38.3 & 27.1 & - \\
R3$^2$                   & 47.3 & 53.7 & 29.1 & 37.5 \\ 
DS-QA$^3$                & 48.7 & 56.3 & 28.7 & 36.6 \\
Re-Ranker$^4$            & 50.6 & 57.3 & - & - \\
MINIMAL$^5$              & - & - & 34.7 & 42.5 \\
Multi-Step$^6$           & 51.9 & 61.7  & 31.9 & 39.2 \\
S-Norm$^7$               & 61.3 & 67.2  & - & - \\
HAS-QA$^8$               & 63.6 & 68.9  & - & - \\
BERTserini$^9$			 & - & - & 38.6 & 46.1 \\	
\midrule
\reqabase            & 64.1 & 69.8 & 40.1 & 48.4 \\
\reqalarge           & \textbf{65.5} & \textbf{71.2} & \textbf{41.9} & \textbf{50.2} \\
\bottomrule
\end{tabular}
\caption{\label{table:tri-unfil} Results on TriviaQA-unfiltered test set and SQuAD-open dev set: \citet{chen2017reading}$^1$, \citet{wang2018r3}$^2$, \citet{lin2018denoising}$^3$, \citet{wang2017evidence}$^4$, \citet{min2018efficient}$^5$, \citet{das2018multi}$^6$, \citet{clark2017simple}$^7$, \citet{pang2018has}$^8$ and \citet{yang2019end}$^9$.}
\end{center}
\end{table}

\begin{table}
\begin{center}
\small
\begin{tabular}{l|cccccc}
\toprule
\multirow{2}*{ Model } & \multicolumn{2}{c}{ TriviaQA-Wikipedia } & \multicolumn{2}{c}{ SQuAD-document } \\
 & F1 & Speed & F1 & Speed \\ 
\midrule
RE$^3$QA              & 72.68 & 4.62 & 84.81 & 3.76 \\
\bertpipe             & 71.13 & 2.05 & 83.65 & 1.78 \\
\bertpipeplus         & 71.59 & 2.08 & 84.04 & 1.82 \\
\bottomrule
\end{tabular}
\caption{\label{table:pip-triviaqa} Comparison between our approach and the pipelined method. ``Speed'' denotes the number of instances processed per second during inference.}
\end{center}
\end{table}

\begin{figure*}
\center
\begin{minipage}[b]{.48\textwidth}
\includegraphics[width=\columnwidth]{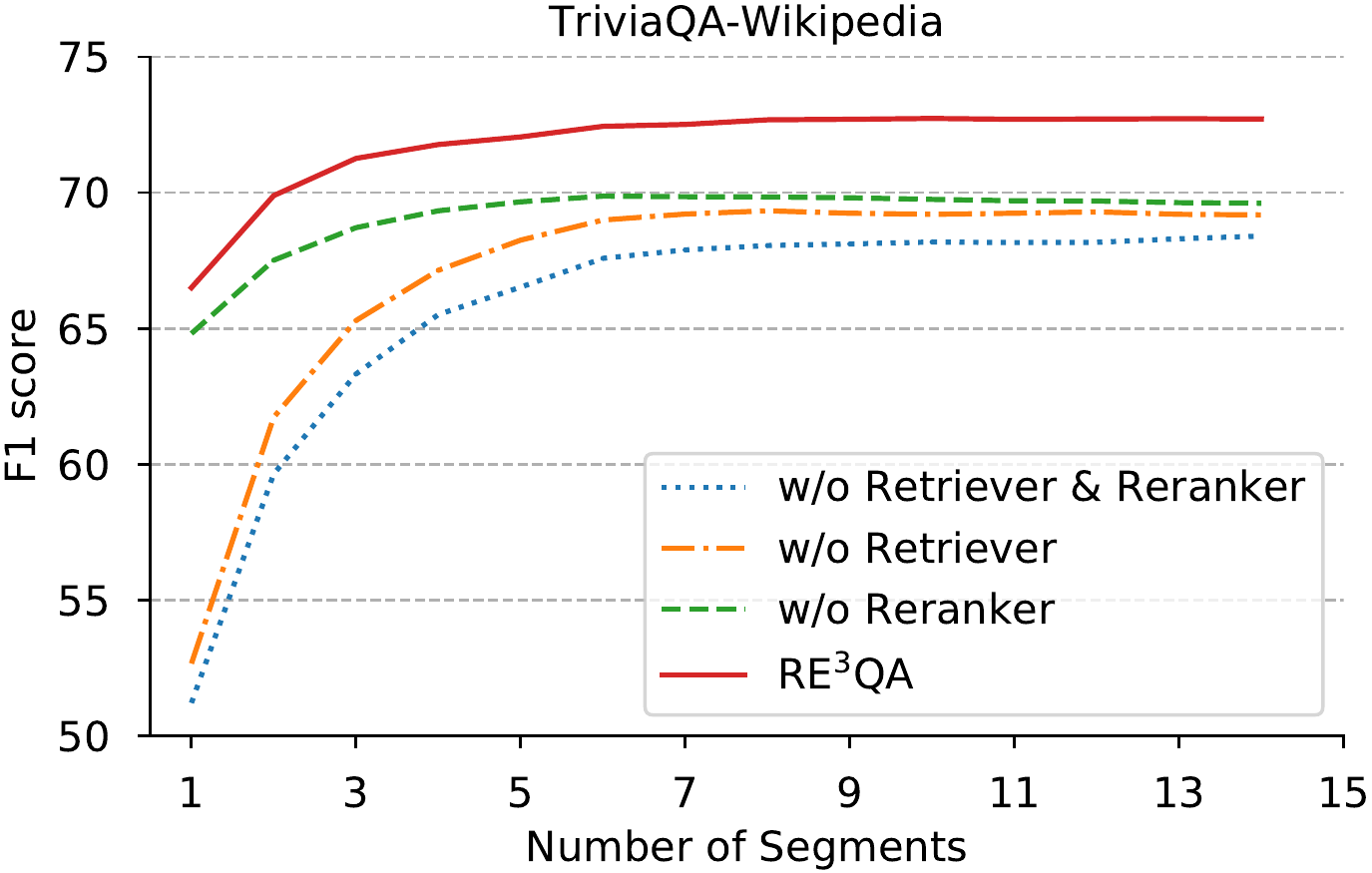}
\end{minipage}
\hfill
\begin{minipage}[b]{.48\textwidth}
\includegraphics[width=\columnwidth]{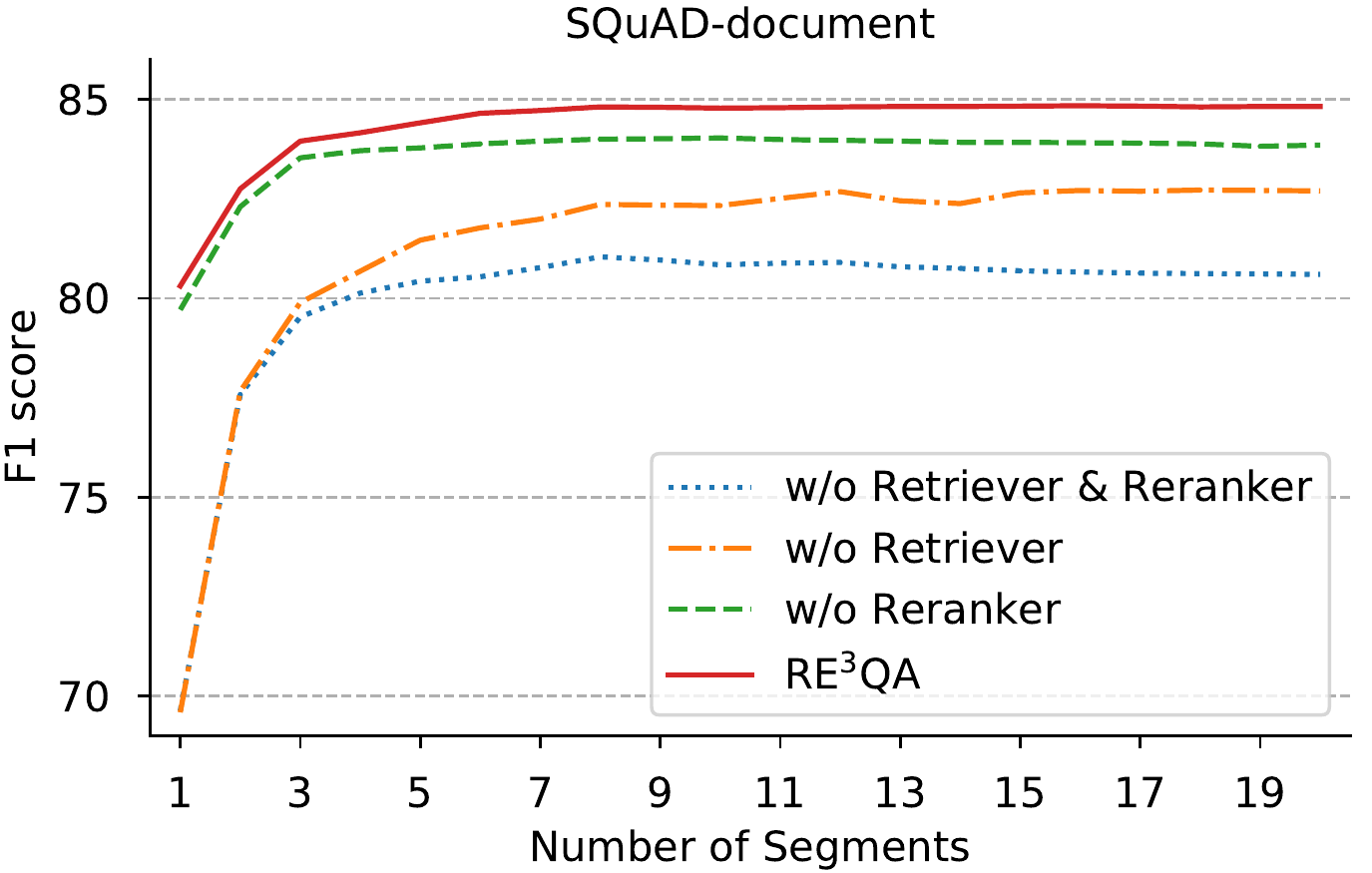}
\end{minipage}
\caption{F1 score on TriviaQA-Wikipedia and SQuAD-document w.r.t different number of retrieved segments.}
\label{fig:f1-num-seg}
\end{figure*}

\begin{table*}
\begin{center}
\begin{tabular}{l|ccccc|ccccc}
\toprule
\multirow{2}*{ $J$ } & \multicolumn{5}{c}{ TriviaQA-Wikipedia } & \multicolumn{5}{c}{ SQuAD-document } \\
 & MAP & Top-3 & Top-5 & F1 & Speed & MAP & Top-3 & Top-5 & F1 & Speed \\ 
\midrule
1              & 67.4 & 81.5 & 87.3 & 69.2 & 5.9 & 39.2 & 47.5 & 66.8 & 54.4 & 5.6 \\
2              & 75.3 & 87.4 & 91.1 & 71.7 & 5.1 & 80.3 & 89.4 & 94.0 & 83.4 & 4.7 \\ 
3              & 77.8 & 88.8 & 91.8 & 72.7 & 4.6 & 88.7 & 94.5 & 96.8 & 84.8 & 3.8 \\
4              & 80.0 & 89.2 & 92.1 & 71.6 & 4.2 & 90.2 & 95.0 & 97.2 & 84.3 & 3.0 \\
5              & 80.6 & 89.6 & 92.3 & 71.7 & 3.5 & 91.0 & 95.6 & 97.6 & 84.3 & 2.3 \\
\bottomrule
\end{tabular}
\caption{\label{table:num_J} Retrieving performance with different number of blocks $J$ used for the early-stopped retriever.}
\end{center}
\end{table*}

\subsection{Model Analysis}
In this section, we analyze our approach by answering the following questions\footnote{The \bertbase model is used by default in this section.}: 
(a) Is end-to-end network superior to the pipeline system?
(b) How does each component contribute to the performance? 
(c) Is early-stopped retriever sufficient for returning high-quality segments? 
(d) How does the reranking loss affect the answer reranker?

\paragraph{Comparison with pipelined method}
First, we compare our approach with the pipelined baselines on TriviaQA-Wikipedia and SQuAD-document development sets in Table \ref{table:pip-triviaqa}.
Our approach outperforms \bertpipe by 1.6/1.2 F1 on two datasets respectively, and is also 2.3/2.1 times faster during inference.
Moreover, RE$^3$QA also beats the \bertpipeplus baseline by 1.1/0.8 F1, even as the parameters of retriever and reader are trained sequentially in \bertpipeplus.
The above results confirm that the end-to-end training can indeed mitigate the context inconsistency problem, perhaps due to multi-task learning and parameter sharing.
Our approach can also obtain inference speedups because of the fact that it avoids re-encoding inputs by sharing contextualized representations.

\paragraph{Ablation study}
To show the effect of each individual component, we plot the F1 curve with respect to different number of retrieved segments in Figure \ref{fig:f1-num-seg}.
We notice that all curves become stable as more text are used, implying that our approach is robust across different amounts of context. 
Next, to evaluate the reranker, we only consider the retrieving and reading scores, and the performance decreases by 2.8/0.8 F1 on two datasets after the reranker is removed.
To ablate the retriever, we select segments based on the TF-IDF distance instead.
The results show that the F1 score reduces by about 3.3 and 2.5 points on two datasets after the ablation.
Removing both the retriever and the reranker performs the worst, which only achieves 68.1/81.0 F1 on two datasets at peak.
The above results suggest that combining retriever, reader, and reranker is crucial for achieving promising performance.

\paragraph{Effect of early-stopped retriever}
We assess whether the early-stopped retriever is sufficient for the segment retrieving task.
Table \ref{table:num_J} details the retrieving and reading results with different number of blocks $J$ being used.
As we can see, the model performs worst but maintains a high speed when $J$ is 1.
As $J$ becomes larger, the retrieving metrices such as MAP, Top-3 and Top-5 significantly increase on both datasets.
On the other hand, the speed continues to decline since more computations have been done during retrieving. 
A $J$ of 6 eventually leads to an out-of-memory issue on both datasets.
As for the F1 score, the model achieves the best result when $J$ reaches 3, and starts to degrade as $J$ continues rising.
We experiment with the \reqalarge model and observe similar results, where the best $J$ is 6.
A likely reason for this observation may be that sharing high-level features with the retriever could disturb the reading prediction.
Therefore, the above results demonstrate that an early-stopped retriever with a relatively small $J$ is able to reach a good trade-off between efficiency and effectiveness.

\paragraph{Effect of answer reranker}
\begin{table}
\begin{center}
\small
\begin{tabular}{l|cccc}
\toprule
\multirow{2}*{ Model } & \multicolumn{2}{c}{ TriviaQA-Wikipedia } & \multicolumn{2}{c}{ SQuAD-document } \\
 & EM & F1 & EM & F1 \\ 
\midrule
RE$^3$QA            & 68.51 & 72.68 & 77.90 & 84.81 \\
w/o NMS             & 68.29 & 72.33 & 77.67 & 84.36 \\
w/o $\mathbf{y}^{hard}$      & 67.36 & 71.87 & 77.26 & 84.17 \\ 
w/o $\mathbf{y}^{soft}$      & 67.76 & 72.29 & 77.04 & 84.05 \\
\bottomrule
\end{tabular}
\caption{\label{table:reran_ablate} Reranking performance with different ablations. 
$\mathbf{y}^{hard}$ and $\mathbf{y}^{soft}$ refer to the two labels used to train the reranker.}
\end{center}
\end{table}
Finally, we run our model under different reranking ablations and report the results in Table \ref{table:reran_ablate}.
As we can see, removing the non-maximum suppression (NMS) algorithm has a negative impact on the performance, suggesting it is necessary to prune highly-overlapped candidate answers before reranking.
Ablating the hard label leads to a drop of 0.81 and 0.64 F1 scores on two datasets respectively, while the F1 drops by 0.39 and 0.76 points after removing the soft label.
This implies that the hard label has a larger impact than the soft label on the TriviaQA dataset, but vice versa on SQuAD.

\section{Conclusion}
We present RE$^3$QA, a unified network that answers questions from multiple documents by conducting the retrieve-read-rerank process.
We design three components for each subtask and show that an end-to-end training strategy can bring in additional benefits.
RE$^3$QA outperforms the pipelined baseline with faster inference speed and achieves state-of-the-art results on four challenging reading comprehension datasets.
Future work will concentrate on designing a fast neural pruner to replace the IR-based pruning component, developing better end-to-end training strategies, and adapting our approach to other datasets such as Natural Questions~\cite{47761}.

\section*{Acknowledgments}
We would like to thank Mandar Joshi for his help with TriviaQA submissions. 
We also thank anonymous reviewers for their thoughtful comments and helpful suggestions.
This work was supported by the National Key Research and Development Program of China
(2018YFB0204300).

\bibliography{sections/reference}
\bibliographystyle{acl_natbib}

\appendix

\begin{table*}[h]
    \begin{center}
    \begin{tabular}{l|ccc}
    \toprule
	\tabincell{l}{\textbf{Question:} Which organisation was founded in Ontario, \\ Canada in 1897 by Adelaide Hoodless?} & \multicolumn{3}{c}{\bf{Scores}} \\ 
	\midrule
	\multicolumn{1}{l|}{\bf{Candidate Answers:}} & \bf{Retrieving} & \bf{Reading} & \bf{Reranking} \\
	\textbf{[1]} Women's Institute & \ensuremath{\bf{0.517}} & \ensuremath{11.226} & \ensuremath{2.093} \\
	\textbf{[2]} Young Women's Christian Association & \ensuremath{0.231} & \ensuremath{11.263} & \ensuremath{\bf{2.299}} \\
	\textbf{[3]} Federated Women's Institutes of Canada & \ensuremath{0.426} & \ensuremath{\bf{11.267}} & \ensuremath{1.742} \\
	\textbf{[4]} Victorian Order of Nurses & \ensuremath{0.360} & \ensuremath{11.139} & \ensuremath{1.837} \\
	\textbf{[5]} National Council of Women & \ensuremath{0.291} & \ensuremath{8.966} & \ensuremath{1.02} \\
	\multicolumn{1}{c|}{\bf{\ldots \ldots}} & \multicolumn{3}{c}{\bf{\ldots \ldots}}\\ 
	\bottomrule                            
	\end{tabular}
	\caption{A sampled case (ID: sfq\_21220) from the TriviaQA-Wikipedia dev set shows that although candidate [2] and candidate [3] get higher reranking and reading scores, the candidate [1] is preferred by the retrieving component and is therefore chosen as the final answer. The ground truth answer is ``\emph{Women's Institute}''.}
	\label{tab:case-study1}
	\end{center}
\end{table*}

\begin{table*}[h]
    \begin{center}
    \begin{tabular}{l|ccc}
    \toprule
	\tabincell{l}{\textbf{Question:} Hong Kong is one of two `special administrative \\ regions' of China; what is the other?} & \multicolumn{3}{c}{\bf{Scores}} \\ 
	\midrule
	\multicolumn{1}{l|}{\bf{Candidate Answers:}} & \bf{Retrieving} & \bf{Reading} & \bf{Reranking} \\
	\textbf{[1]} Macau & \ensuremath{0.195} & \ensuremath{11.067} & \ensuremath{\bf{2.502}} \\
	\textbf{[2]} Kowloon & \ensuremath{\bf{0.346}} & \ensuremath{\bf{11.175}} & \ensuremath{1.795} \\
	\textbf{[3]} Kowloon, and the new territories & \ensuremath{0.346} & \ensuremath{7.941} & \ensuremath{0} \\
	\textbf{[4]} Macau, China & \ensuremath{0.323} & \ensuremath{7.812} & \ensuremath{0} \\
	\textbf{[5]} Taiwan & \ensuremath{0.224} & \ensuremath{5.926} & \ensuremath{0.028} \\
	\multicolumn{1}{c|}{\bf{\ldots \ldots}} & \multicolumn{3}{c}{\bf{\ldots \ldots}}\\ 
	\bottomrule                            
	\end{tabular}
	\caption{A sampled case (ID: sfq\_10640) from the TriviaQA-Wikipedia dev set shows that although the candidate [2] gets higher retrieving and reading scores, the candidate [1] is chosen as the final answer since it has the highest reranking score. The ground truth answer is ``\emph{Macau}''.}
	\label{tab:case-study2}
	\end{center}
\end{table*}

\section{Case Study}
To demonstrate how each component takes effect when predicting the final answer, we conduct some qualitative case studies sampled from the \reqalarge model on the TriviaQA-Wikipedia development set. 
For each question, we list top-5 candidate answers along with their retrieving, reading, and reranking scores.

As shown in Table \ref{tab:case-study1}, we first notice that the top-ranked predictions have highly-relevant semantics and share the same linguistic pattern.
As a result, the top-4 candidates contain very similar reading scores from 11.1 to 11.3, which matches the observations of \citet{clark2017simple}.
A likely reason of this phenomenon is that reading comprehension models are easily fooled by confusing distractors (also referred as adversarial examples mentioned by \citet{Jia17}).
Under such circumstance, it is crucial to perform additional answer verifications to identify the final answer.
In this example, we can see that the retriever becomes the key factor when the reader and reranker are distracted by confusing candidates (e.g., the second and third predictions).
By taking the weighted sum of the three scores, our model eventually predicts the correct answer since the first prediction has the largest retrieving score.

Similar observations can be made in Table \ref{tab:case-study2}.
On the one hand, despite the confusing candidate ``\emph{Kowloon}'' has the highest retrieving and reading scores, the reranker assigns a larger confidence on the candidate ``\emph{Macau}''.
As a result, ``\emph{Macau}'' is chosen as the final answer.
On the other hand, we find that the reranking scores of some candidates (e.g., the third and fourth predictions) are zero.
This is due to the span-level non-maximum suppression algorithm, where redundant spans such as ``\emph{Macau, China}'' will be pruned before the reranking step.
Therefore, the final weighted-sum scores of these candidates will be significantly lower than the top predictions, which is beneficial for filtering distractors out.

\end{document}